\def\year{2020}\relax
\documentclass[letterpaper]{article} 
\usepackage{aaai20}  
\usepackage{times}  
\usepackage{helvet} 
\usepackage{courier}  
\usepackage[hyphens]{url}  
\usepackage{graphicx} 
\usepackage{graphicx}  
\usepackage{amsmath}
\usepackage{caption}
\usepackage{subcaption}
\usepackage[table]{xcolor}

\title{GlobalTrack: A Simple and Strong Baseline for Long-term Tracking}
\author{Lianghua Huang\textsuperscript{\rm 1,2}, Xin Zhao\textsuperscript{\rm 1,2}\thanks{Corresponding author}, Kaiqi Huang\textsuperscript{\rm 1,2,3}\\
\textsuperscript{\rm 1}CRISE, Institute of Automation, Chinese Academy of Sciences, Beijing, China\\
\textsuperscript{\rm 2}University of Chinese Academy of Sciences, Beijing, China\\
\textsuperscript{\rm 3}CAS Center for Excellence in Brain Science and Intelligence Technology, Beijing, China\\
huanglianghua2017@ia.ac.cn, \{xzhao, kqhuang\}@nlpr.ia.ac.cn}
 
\begin{document}

\maketitle

\begin{abstract}
A key capability of a long-term tracker is to search for targets in very large areas (typically the entire image) to handle possible target absences or tracking failures. However, currently there is a lack of such a strong baseline for global instance search. In this work, we aim to bridge this gap. Specifically, we propose GlobalTrack, a pure global instance search based tracker that makes no assumption on the temporal consistency of the target's positions and scales. GlobalTrack is developed based on two-stage object detectors, and it is able to perform full-image and multi-scale search of arbitrary instances with only a single query as the guide. We further propose a cross-query loss to improve the robustness of our approach against distractors. With no online learning, no punishment on position or scale changes, no scale smoothing and no trajectory refinement, our pure global instance search based tracker achieves comparable, sometimes much better performance on four large-scale tracking benchmarks (i.e., 52.1\% AUC on LaSOT, 63.8\% success rate on TLP, 60.3\% MaxGM on OxUvA and 75.4\% normalized precision on TrackingNet), compared to state-of-the-art approaches that typically require complex post-processing. More importantly, our tracker runs without cumulative errors, i.e., any type of temporary tracking failures will not affect its performance on future frames, making it ideal for long-term tracking. We hope this work will be a strong baseline for long-term tracking and will stimulate future works in this area. Code is available at \url{https://github.com/huanglianghua/GlobalTrack}.
\end{abstract}

\noindent Given an arbitrary, user-specified target in the first frame, the task of visual tracking is to locate it continuously in successive frames. Visual tracking has been widely used in many fields such as surveillance, augmented reality, robotics and video editing. Over the past decades, significant progress has been made in this area~\cite{otb2015,oxuva2018,got2019}.

\begin{figure}[t]
  \centering
  \includegraphics[width=0.49\textwidth]{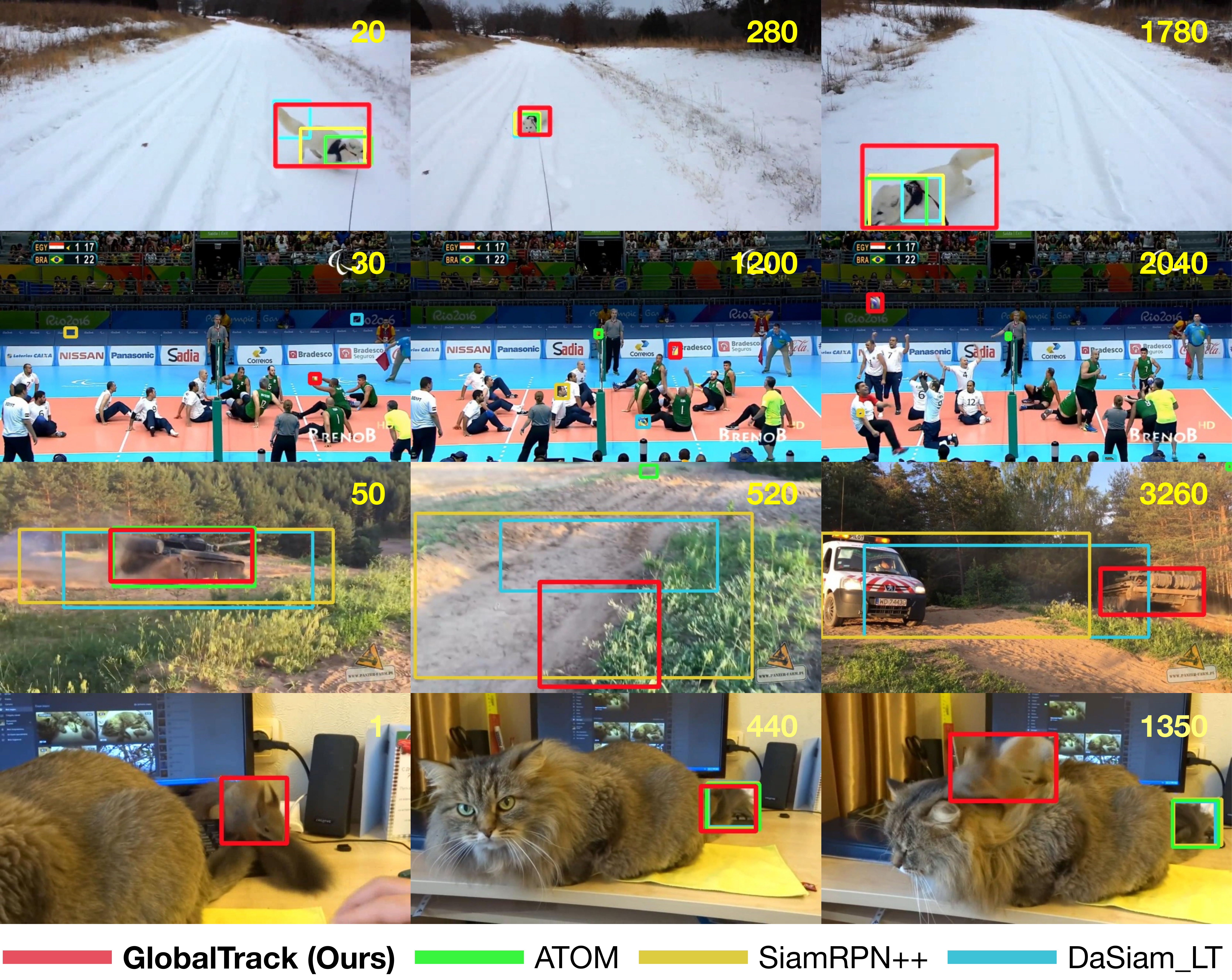}
  \caption{
    A comparison of our approach with state-of-the-art trackers. ATOM~\cite{atom2019}, SiamRPN++~\cite{siamrpnpp2019} and DaSiam\_LT~\cite{dasiamrpn2018} work under the temporal consistency assumption that the target's state changes smoothly.
    However, as the figure shows, such assumption does not necessarily hold. Under abrupt motion and temporary target absence, these trackers cannot locate the targets for long.
    Our approach GlobalTrack, a pure global instance search based tracker, successfully handles these challenges and provides robust tracking results.}
  \label{fig:1}
\end{figure}

Most existing trackers work under a strong temporal consistency assumption that the target's position and scale change smoothly. These approaches typically search within a small window for the target and impose penalties on large position and scale variations to limit the prediction space~\cite{siamfc2016,siamrpn2018,iccv19,atom2019}.
However, various real-world challenges can break such assumption and cause these approaches to fail.
Figure~\ref{fig:1} shows typical failure cases of some state-of-the-art trackers, where under abrupt position and scale changes, target absences and temporary tracking failures, none of these approaches is able to consistently locate the targets for long.

\begin{figure*}[t]
  \centering
  \includegraphics[width=0.96\textwidth]{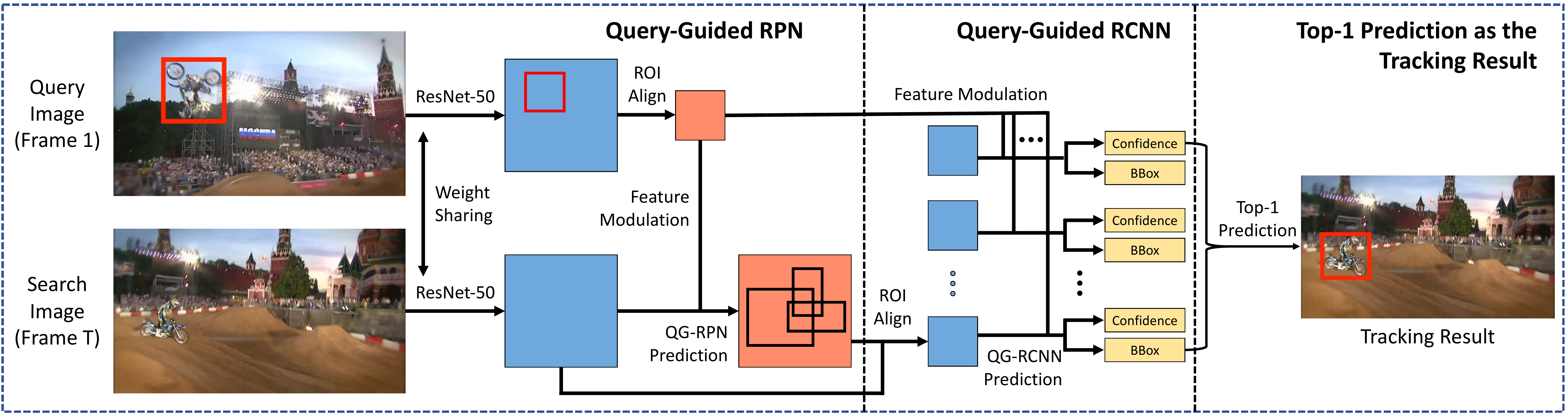}
  \caption{The overall architecture of GlobalTrack. The architecture consists of two submodules: a Query-Guided RPN (QG-RPN) for generating query-specific proposals, and a Query-Guided RCNN (QG-RCNN) for discriminating the proposals and producing the final predictions. In the feature modulation of QG-RPN and QG-RCNN, we encode the correlation between the query and search image features in the backbone and ROI outputs, so as to guide the detector to locate query-specific instances. During tracking, we use the first frame as the query and simply take the top-1 prediction in a frame as the tracking result.}
  \label{fig:architecture}
\end{figure*}

In this paper, we propose a baseline tracker that handles these challenges using pure global instance search. The key idea is to remove the locality assumption and enable the tracker to search for the target at arbitrary positions and scales, thereby avoiding cumulative errors during tracking.
We build such global instance search algorithm based on object detection models, since they are able to perform full-image and multi-scale search of arbitrary sized objects.

Specifically, we propose GlobalTrack, a full-image visual tracker, inspired by the two-stage object detector Faster-RCNN~\cite{fasterrcnn2015}. Similar to Faster-RCNN, GlobalTrack consists of two submodules: a query-guided region proposal network (QG-RPN) for generating query-specific object candidates, and a query-guided region convolutional neural network (QG-RCNN) for classifying the candidates and producing the final predictions. The overall architecture of GlobalTrack is visualized in Figure~\ref{fig:architecture}. In the feature modulation parts of QG-RPN and QG-RCNN, we encode the correlation between the query and search image features in the backbone and ROI outputs, thereby directing the detector to locate query-specific instances.

During tracking, we take the annotated first frame as the query and search independently in each of the rest frames for the target. We simply take the top-1 prediction of QG-RCNN as the tracking result, without using any further post-processing. While adding additional processes, such as trajectory smoothing, may improve the performance of our approach, we prefer to keep the current model simple and straightforward.
In the training phase, we sample frame pairs from video datasets to optimize the model, utilizing the same classification and localization losses as in Faster-RCNN.
We further propose a cross-query loss to improve the robustness of GlobalTrack against instance-level distractors, which averages the losses over different queries on a same image to force the model to learn the strong dependency between the queries and the prediction results.

\begin{figure}[t]
  \centering
  \includegraphics[width=0.46\textwidth]{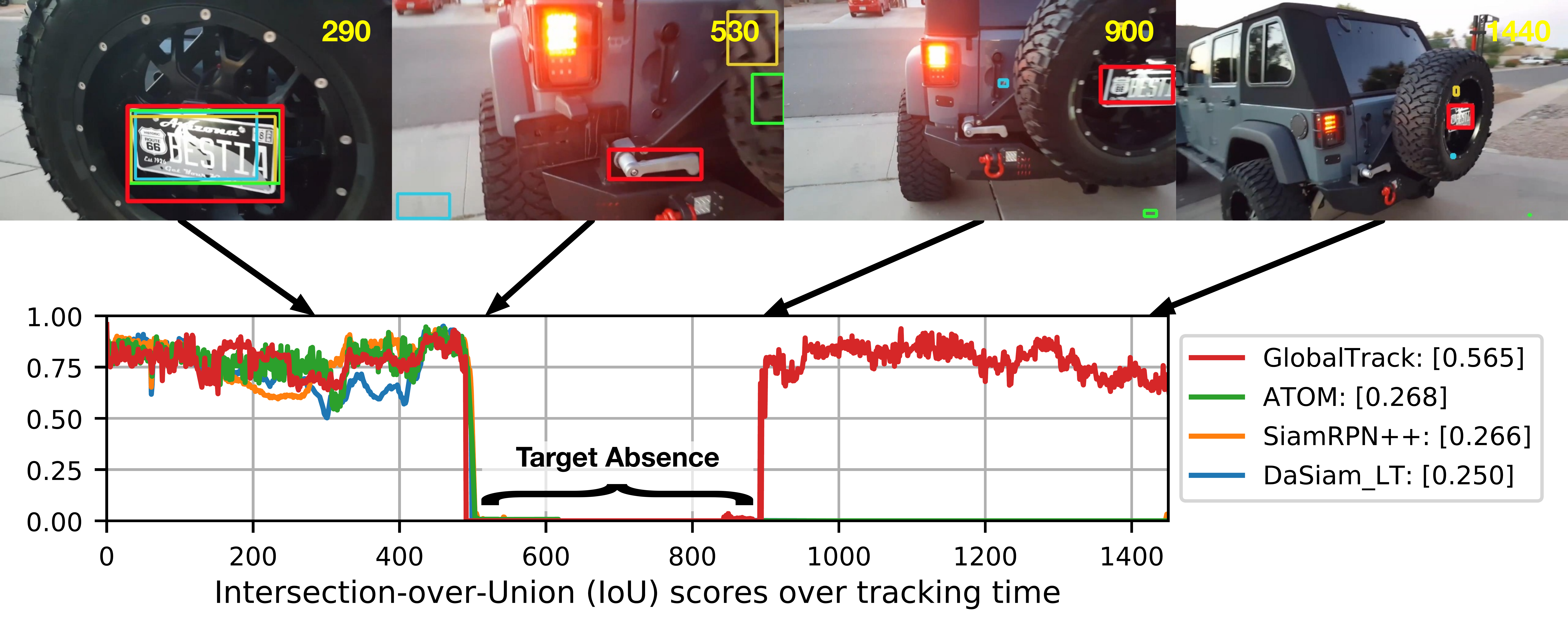}
  \caption{A long-term tracking example where the target undergoes a period of ($\sim$400 frames) absence. All compared methods fail to relocate the target after the temporary absence, while GlobalTrack relocates it immediately as it reappears, and achieves a much higher average IoU score. The example shows the advantages of GlobalTrack -- since it makes no locality assumption and searches globally, its performance in a frame is not affected by its previous failures.}
  \label{fig:2}
\end{figure}

We verify the performance of our approach on four large scale tracking benchmarks: LaSOT~\cite{lasot2019}, TrackingNet~\cite{trackingnet2018}, TLP~\cite{tlp2018} and OxUvA~\cite{oxuva2018}, where LaSOT, OxUvA and TLP are long-term tracking benchmarks with average video lengths of 2500, 4260 and 13529 frames, respectively.
GlobalTrack shows impressive performance on these datasets, compared to state-of-the-art approaches that typically require complex post-processing. For example, GlobalTrack outperforms SiamRPN++~\cite{siamrpnpp2019} and ATOM~\cite{atom2019} by achieving an AUC (area-under-curve) of 52.1\% on LaSOT benchmark; it also achieves an large-margin absolute gain of 11.1\% on TLP benchmark, over the previous best long-term tracker SPLT~\cite{splt2019}.
More importantly, since our approach imposes no assumption on temporal consistency, its performance in one frame is completely independent of previous tracking failures, enabling our model to track very long videos without suffering from cumulative errors.
Figure~\ref{fig:2} shows an example where the target disappears temporarily for around 400 frames. The compared approaches fail to relocate the target as it reappears, while GlobalTrack catches the target immediately as it shows up, showing the advantage of our approach in long-term tracking scenarios.

\section{Related Work}

Long-term tracking refers to the task of continuously locating an arbitrary target in a relatively long video, where the target may temporarily disappear~\cite{oxuva2018}.
A key challenge of long-term tracking is to retrieve the target after a period of target absence or tracking failures.
To our knowledge, only limited works focus on this task~\cite{tld2012,lct2015,bmvc19,icip19}.
Representive approaches include TLD~\cite{tld2012}, SPL~\cite{spl2013}, LCT~\cite{lct2015}, EBT~\cite{ebt2016} and DaSiam\_LT~\cite{dasiamrpn2018}.
TLD, SPL, LCT and EBT are capable of performing full-image search, which is an important ability, especially when the target can disappear. TLD, SPL and LCT searches globally for the target when they detect tracking failures, while EBT always perform full-image search.
DaSiam\_LT~\cite{dasiamrpn2018} is a variant of DaSiamRPN~\cite{dasiamrpn2018} for long-term tracking, which expands the search area as it detects tracking failures.
Similar to these trackers, our approach GlobalTrack can perform full-image search for the target. Differently, we use no online learning and impose no constraints on temporal consistency of the target's location or scale to avoid cumulative errors.

In terms of the tracking framework, our approach share some similarities with SiamRPN~\cite{siamrpn2018} and ATOM~\cite{atom2019}, where they all use shared backbones for extracting features of the query and search images, and they use correlation to encode the relationship between them.
However, our work has significant differences with these methods: (1) Our approach performs global search for the target in all frames, without relying on any locality assumption, while ATOM and SiamRPN search only locally and they use complex post-processing to impose constraints on the target's location and scale changes; (2) We reuse the RPN and RCNN heads of Faster-RCNN and apply them on the modulated features to generate proposals and final predictions. However, SiamRPN uses the query to generate classification and localization weights, while ATOM learns a hand-crafted classifier from scratch during tracking.

\section{Our Approach}

In this work, we propose GlobalTrack, a pure global instance search based tracker consisting of two components: a Query-Guided RPN (QG-RPN) for adaptively generating proposals that are specific to a certain query, and a Query-Guided RCNN (QG-RCNN) for classifying the proposals and generating final predictions.
The overall architecture of GlobalTrack is shown in Figure~\ref{fig:architecture}.
In the feature modulation parts of QG-RPN and QG-RCNN, we encode the correlation between the query and search image features in the outputs of the backbone and ROI layers, thereby allowing the model to learn the strong dependency between the queries and the expected prediction results.

In the following, we first introduce the QG-RPN and QG-RCNN in details. Then we present the cross-query loss used in our approach. Finally, we introduce the offline training and online tracking processes of our approach.

\subsection{Query-guided RPN}

Region Proposal Networks (RPNs) are widely used in two-stage object detectors to generate class-agnostic object candidates of pre-defined classes and narrow the search scope down.
However, these RPNs are general-purposed; while in tracking, we are only interested in candidates of specific targets.
We propose a Query-Guided RPN (QG-RPN) to achieve this, where the key idea is to use correlation to encode the query information in backbone features.

Specifically, let $z \in \mathrm{R}^{k\times k\times c}$ denotes the ROI (Region-Of-Interest) features of the query instance, and $x\in \mathrm{R}^{h\times w\times c}$ represents the search image features, where $h, w$ and $k$ represent feature sizes; we aim to obtain an $\hat{x}\in \mathrm{R}^{h\times w\times c}$ that encodes the correlation between $z$ and $x$:
\begin{eqnarray}\label{eq:rpn_modulation}
  \hat{x} = g_{qg\_rpn}(z, x) = f_{out} (f_x(x) \otimes f_z(z)).
\end{eqnarray}
Here $\otimes$ denotes the convolution operator, $f_z(z)$ converts $z$ to a convolutional kernel that is applied on the projected features $f_x(x)$ to generate the correlation between $z$ and $x$. $f_{out}$ ensures the output $\hat{x}$ to have the same size as $x$.
For simplicity, in our approach, we define $f_z$ to be a $k\times k$ convolution layer with zero padding that converts $z$ to a $1\times 1$ convolutional kernel, $f_x$ to be a $3\times 3$ convolution layer with 1 pixel padding, while $f_{out}$ to be a $1\times 1$ convolution layer that converts the channel number back to $c$.
We use no normalization and no activation in these projections.

Since $\hat{x}$ retains the size of $x$, we directly reuse the modules of RPN and perform its subsequent processes (classification, localization, filtering, etc.) to generate the proposals. We use the same losses of RPN for training QG-RPN, where the classification and localization losses $L_{cls}$ and $L_{loc}$ are binary cross-entropy and smooth L1, respectively~\cite{fasterrcnn2015}. The total loss of QG-RPN is:
\begin{eqnarray}\label{eq:rpn_loss}
  L_{qg\_rpn} (z, x) = L_{rpn} (\hat{x}) = \frac{1}{N_{cls}} \sum_i L_{cls} (p_i, p_i^*) + \nonumber \\
             \lambda \frac{1}{N_{loc}} \sum_i p_i^* L_{loc} (s_i, s_i^*)
\end{eqnarray}
where $p_i$ and $s_i$ are the predicted score and location of the $i$th proposal, while $p_i^*$ and $s_i^*$ are groundtruths. $\lambda$ is a weight for balancing the classification and localization losses.

\subsection{Query-guided RCNN}

With proposals generated by QG-RPN, in the second stage, we need to refine the predictions of their labels and bounding boxes according to the ROI features. Note such process is query-specific, since apparently different queries correspond to different groundtruths.
We propose Query-Guided RCNN (QG-RCNN) for the classification and bounding box refinement of these proposals.
Specifically, given the ROI features of the query $z\in \mathrm{R}^{k\times k\times c}$ and the $i$th proposal $x_i\in \mathrm{R}^{k\times k\times c}$, we perform feature modulation to encode their correlation:
\begin{eqnarray}\label{eq:rcnn_modulation}
  \hat{x}_i = g_{qg\_rcnn}(z, x_i) = h_{out} (h_x(x_i) \odot h_z(z)),
\end{eqnarray}
where $\odot$ denotes the Hadamard production, $h_x$ and $h_z$ are feature projections for $x_i$ and $z$, respectively, while $h_{out}$ generate output features $\hat{x}$ and ensures that it retains the size of $x_i$.
In our approach, we simply set $h_x$ and $h_z$ to $3\times 3$ convolution layers with 1 pixel padding, and $h_{out}$ to a $1\times 1$ convolution layer with output channel number being $c$.

After obtaining the modulated features $\hat{x}_i$, we continue with the traditional RCNN processes and perform classification and localization on the proposals to obtain the final predictions. During training, similar to QG-RPN, we use binary cross-entropy and smooth L1 as the classification and localization losses to optimize our model. The total loss of QG-RCNN is formulated as:
\begin{eqnarray}\label{eq:rcnn_loss_total}
  L_{qg\_rcnn} (z, x) = \frac{1}{N_{prop}} \sum_i L_{rcnn} (\hat{x}_i),
\end{eqnarray}
where $N_{prop}$ is the proposal number and
\begin{eqnarray}\label{eq:rcnn_loss}
  L_{rcnn} (\hat{x}_i) = L_{cls} (p_i, p_i^*) + \lambda p_i^* L_{loc} (s_i, s_i^*).
\end{eqnarray}
Here the $p_i$ and $s_i$ are the estimated confidence and location (center and scale offsets), while $p_i^*$ and $s_i^*$ are groundtruths. $\lambda$ is a weight for balancing different losses.

\subsection{Cross-query Loss}

To improve the discriminative ability of our approach against instance-level distractors, we propose the cross-query loss. The key idea is to enhance the awareness of our model on the relationship between queries and prediction outputs, by searching a same image using different queries and average their prediction losses.
Formally speaking, given a pair of images with $M$ co-exist instances $\{1, 2, \cdots, M\}$, we can construct $M$ query-search image pairs, and thereby calculating $M$ QG-RPN and QG-RCNN losses. We average the losses over these queries to obtain the final loss on a pair of images:
\begin{eqnarray}\label{eq:cql_loss}
  L_{cql} = \frac{1}{M} \sum_{k=1}^M L (z_k, x),
\end{eqnarray}
where
\begin{eqnarray}\label{eq:single_loss}
  L(z_k, x) = L_{qg\_rpn} (z_k, x) + L_{qg\_rcnn} (z_k, x).
\end{eqnarray}
Note the calculation of $L_{cql}$ is very efficient, since the $M$ query-search image pairs share most computation in backbone feature extraction.

\subsection{Offline Training}

In the training phase, we randomly sample frame pairs from training videos to construct training data. For each frame pair $I_z$ and $I_x$, we find the $M$ instances that co-exist in both frames, then we construct $M$ query-search image pairs accordingly. We run one forward pass of backbone network on $I_z$ and $I_x$ to obtain their feature maps, then we perform $M$ times of feature modulation using Eq.~\eqref{eq:rpn_modulation} and Eq.~\eqref{eq:rcnn_modulation}. The modulated features are fed into QG-RPN and QG-RCNN to obtain the query-specific predictions. The loss of the frame pair is then evaluated using Eq.~\eqref{eq:cql_loss}, which is an average of losses over all available queries. The data sampling and loss evaluation are performed for several iterations, and we use the standard stochastic gradient descent (SGD) algorithm to optimize our model.

\subsection{Online Tracking}

The tracking process of our approach is extremely simple. In the first frame, we initialize the query using the user-specified annotation. Then the query is fixed throughout the tracking process with no updating. In a new tracking frame, we run a forward pass of QG-RPN and QG-RCNN with the query and current image as inputs. Finally, we directly take the top-1 prediction of QG-RCNN in this frame as the tracking result. No further post-processing is used.
Although adding some post-processing, such as casting penalization on large state variations or performing trajectory refinement, may improve the performance of our approach, we prefer to keep the current model simple and straightforward, and leave a more adaptive tracking model to our future works.

\begin{figure}[t]
  \centering
  \includegraphics[width=0.4\textwidth]{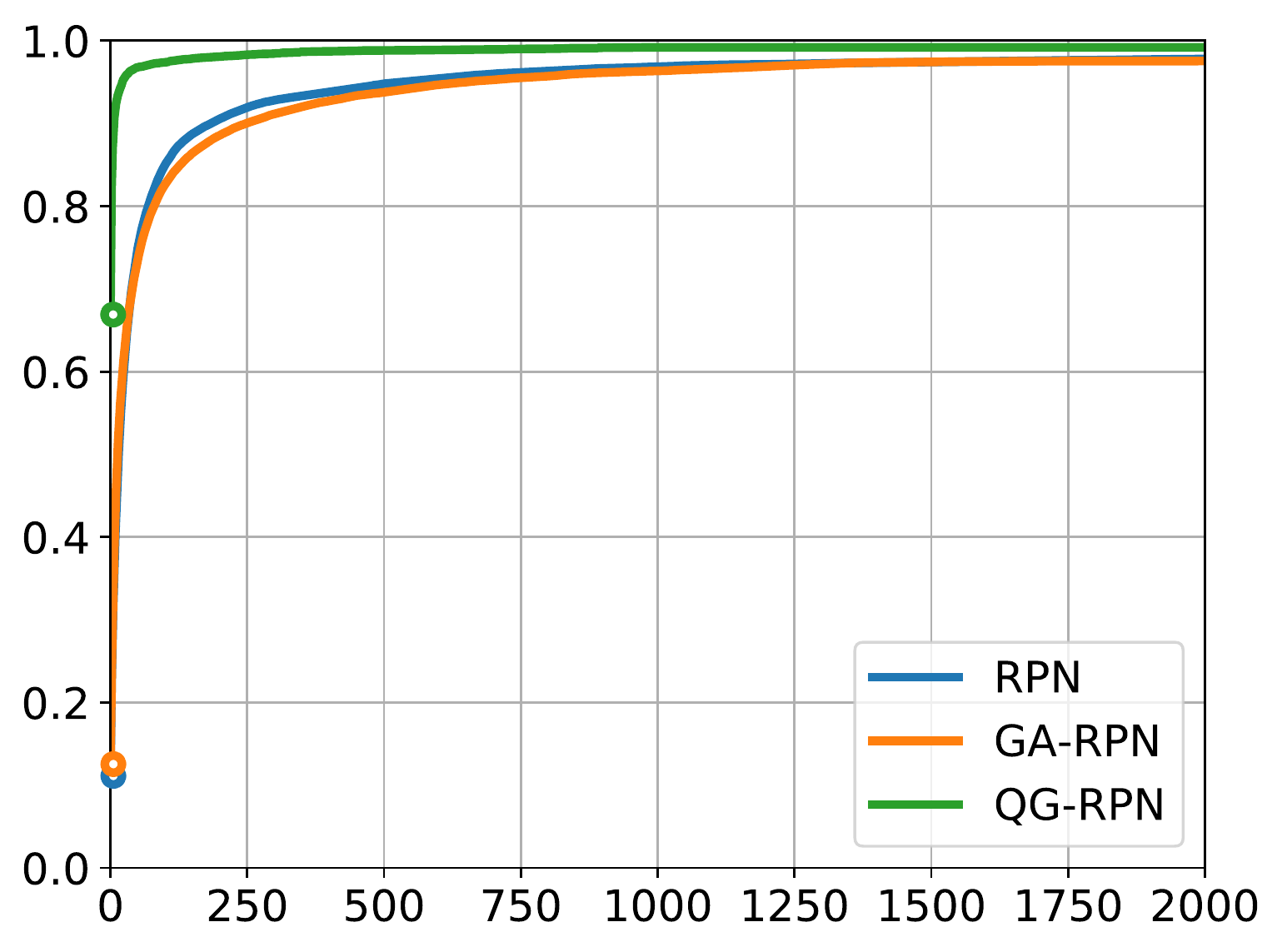}
  \caption{AR@k plots of different region proposal networks (RPNs). Compared to general purposed proposal networks RPN and GA-RPN, the proposed query-guided RPN consistently show a much higher recall.}
  \label{fig:ar_k_rpns}
\end{figure}
\begin{table}[t]
   \footnotesize
   \begin{center}
       \caption{AR@k of different region proposal networks (RPNs). QG-RPN shows high recall even with very few proposals. Moreover, with all proposals available, the average recall of QG-RPN still surpasses the compared approaches.}
       \label{tab:ar_k_rpns}
       \begin{tabular}{
            >{\raggedright\arraybackslash} m{1.25cm}
            >{\centering\arraybackslash} m{0.7cm}
            >{\centering\arraybackslash} m{0.8cm}
            >{\centering\arraybackslash} m{0.95cm}
            >{\centering\arraybackslash} m{0.95cm}
            >{\centering\arraybackslash} m{1.25cm}
            }
            \hline
            (\%)          & AR@1 & AR@10 & AR@100 & AR@500 & AR@2000 \\
            \hline
            RPN           & 11.3 & 40.5 & 84.9 & 94.8 & 97.8 \\
            GA-RPN        & 12.9 & 43.3 & 82.6 & 93.8 & 97.5 \\
            \hline
            {\bf QG-RPN}  & {\bf 67.1} & {\bf 91.9} & {\bf 97.4} & {\bf 98.8} & {\bf 99.2} \\
            \hline
       \end{tabular}
   \end{center}
\end{table}

\section{Experiments}

To verify the effectiveness of our approach, we conduct evaluation on four large-scale tracking benchmarks: LaSOT~\cite{lasot2019}, TrackingNet~\cite{trackingnet2018}, TLP~\cite{tlp2018} and OxUvA~\cite{oxuva2018}, where their test sets contain 280, 511, 180 and 166 videos, respectively. LaSOT, TLP and OxUvA are long-term tracking benchmarks with average video lengths of 2500, 13529 and 4260 frames, respectively. We compare our overall results with state-of-the-art approaches in this section. We also carry out experiments to analyze the effectiveness of our individual components.

\subsection{Implementation Details}

\subsubsection{Parameters}
We use Faster-RCNN with ResNet-50 backbone~\cite{fasterrcnn2015} as our base model for constructing query-guided RCNN. The channel number of the backbone features is $c=256$. We set the output channel number of $f_x, f_z$ and $h_x, h_z$ to $c'=256$ as well. In this way, the $f_z$ in effect converts the query features $z$ to a $1\times 1$ depth-wise convolutional kernel, which is more efficient in computation.
We employ ROI align~\cite{maskrcnn2017} to extract ROI features, where the output feature size is set to $k=7$.
We normalize the size of each input image so that its longer edge is no larger than $1333$ pixels, while its shorter edge is no larger than $800$ pixels.
The QG-RPN generates 2000 proposals for calculating its losses, of which 512 proposals are sampled to be fed into QG-RCNN.
The localization loss weight in Eq.~\eqref{eq:rpn_loss} and Eq.~\eqref{eq:rcnn_loss} is set to $\lambda = 1$.

\begin{table}[t]
   \footnotesize
   \begin{center}
       \caption{Numerical comparison of QG-RPN and QG-RCNN in AR@k. QG-RCNN outperforms QG-RPN in AR@1 by around 9.5\%. As more candidates are introduced, QG-RPN achieves higher average recall than QG-RCNN by around 5\%. As the number of candidates gets larger, the average recall of QG-RPN and QG-RCNN become close.}
       \label{tab:ar_k_rpn_rcnn}
       \begin{tabular}{
            >{\raggedright\arraybackslash} m{1.45cm}
            >{\centering\arraybackslash} m{0.75cm}
            >{\centering\arraybackslash} m{0.75cm}
            >{\centering\arraybackslash} m{0.85cm}
            >{\centering\arraybackslash} m{0.92cm}
            >{\centering\arraybackslash} m{1.12cm}
            }
            \hline
            (\%)          & AR@1 & AR@5 & AR@10 & AR@100 & AR@512 \\
            \hline
            QG-RPN        & 67.1 & {\bf 87.1} & {\bf 91.9} & {\bf 97.4} & 98.8 \\
            QG-RCNN       & {\bf 76.6} & 82.5 & 85.9 & 97.4 & {\bf 98.9} \\
            \hline
       \end{tabular}
   \end{center}
\end{table}
\begin{table}[t]
   \footnotesize
   \begin{center}
       \caption{Comparison of GlobalTrack trained using cross-query loss and single-query loss. The performance is evaluated on the test set of LaSOT. The results show that the model trained with cross-query loss consistently outperforms that trained with single-query loss in all three metrics.}
       \label{tab:loss_comparison}
       \begin{tabular}{
            >{\raggedright\arraybackslash} m{2.3cm}
            >{\centering\arraybackslash} m{1.1cm}
            >{\centering\arraybackslash} m{2.2cm}
            >{\centering\arraybackslash} m{1.1cm}
            }
            \hline
            (\%)                & Precision & Norm. Precision & Success \\
            \hline
            Single-query loss   & 49.3 & 55.7 & 49.5 \\
            Cross-query loss    & {\bf 52.7} & {\bf 59.9} & {\bf 52.1} \\
            \hline
       \end{tabular}
   \end{center}
\end{table}

\subsubsection{Training Data}
We use a combination of COCO~\cite{coco2014}, GOT-10k~\cite{got2019} and LaSOT~\cite{lasot2019} datasets for training our model, where the sampling probabilities of the three datasets are 0.4, 0.4 and 0.2, respectively. COCO is an image object detection dataset containing over 118 thousand images, which belong to 80 object classes. GOT-10k and LaSOT are visual tracking datasets, where GOT-10k consists of 10,000 videos belonging to 563 object classes, while LaSOT consists of 1,400 videos belonging to 70 object classes.
For COCO dataset, we randomly sample an image and perform data augmentation on it to generate an image pair; while for GOT-10k and LaSOT datasets, we directly sample frame pairs from videos.
We use random horizontal flipping and color jitter to augment the image pairs and to enrich our training data.

\begin{figure}[t]
  \centering
  \includegraphics[width=0.4\textwidth]{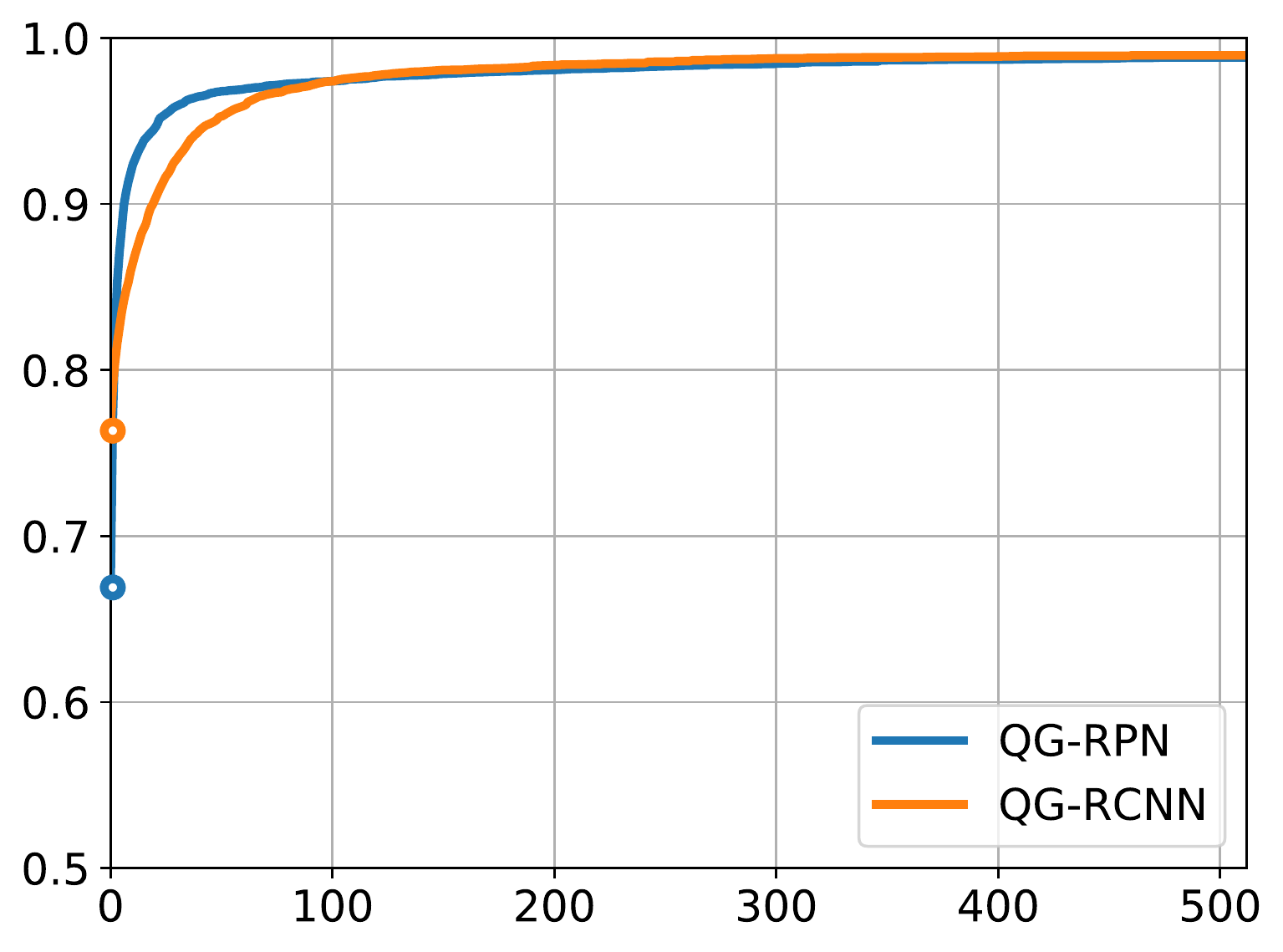}
  \caption{AR@k plots of QG-RPN and QG-RCNN. QG-RCNN achieves much higher AR@1 (equivalent to the top-1 accuracy) than QG-RPN, while its average recall is soon surpassed by QG-RPN as more candidates are introduced. The results demonstrate the different preferences of QG-RPN and QG-RCNN in recall and accuracy.}
  \label{fig:ar_k_rpn_rcnn}
\end{figure}

\subsubsection{Optimization}
We use stochastic gradient descent with a batch size of 4 pairs to train our model. The momentum and weight decay are set to 0.9 and $1\times 10^{-4}$, respectively.
The backbone of our model is initialized from Faster-RCNN pretained on COCO dataset. We fix the batch normalization parameters during training while allowing all other parameters trainable. We train our model for 12 epochs on COCO and another 12 epochs on a combination of COCO, GOT-10k and LaSOT datasets, as described in the previous subsection. The initial learning rate is set to $0.01$, and it decays with a factor of $0.1$ at epoch 8 and 11.
Our approach is implemented in Python, using PyTorch. The training process takes about 16 hours on four GTX TitanX GPUs, while the online tracking runs at around 6 fps on a single gpu.

\begin{figure*}[t]
  \centering
  \includegraphics[width=0.75\textwidth]{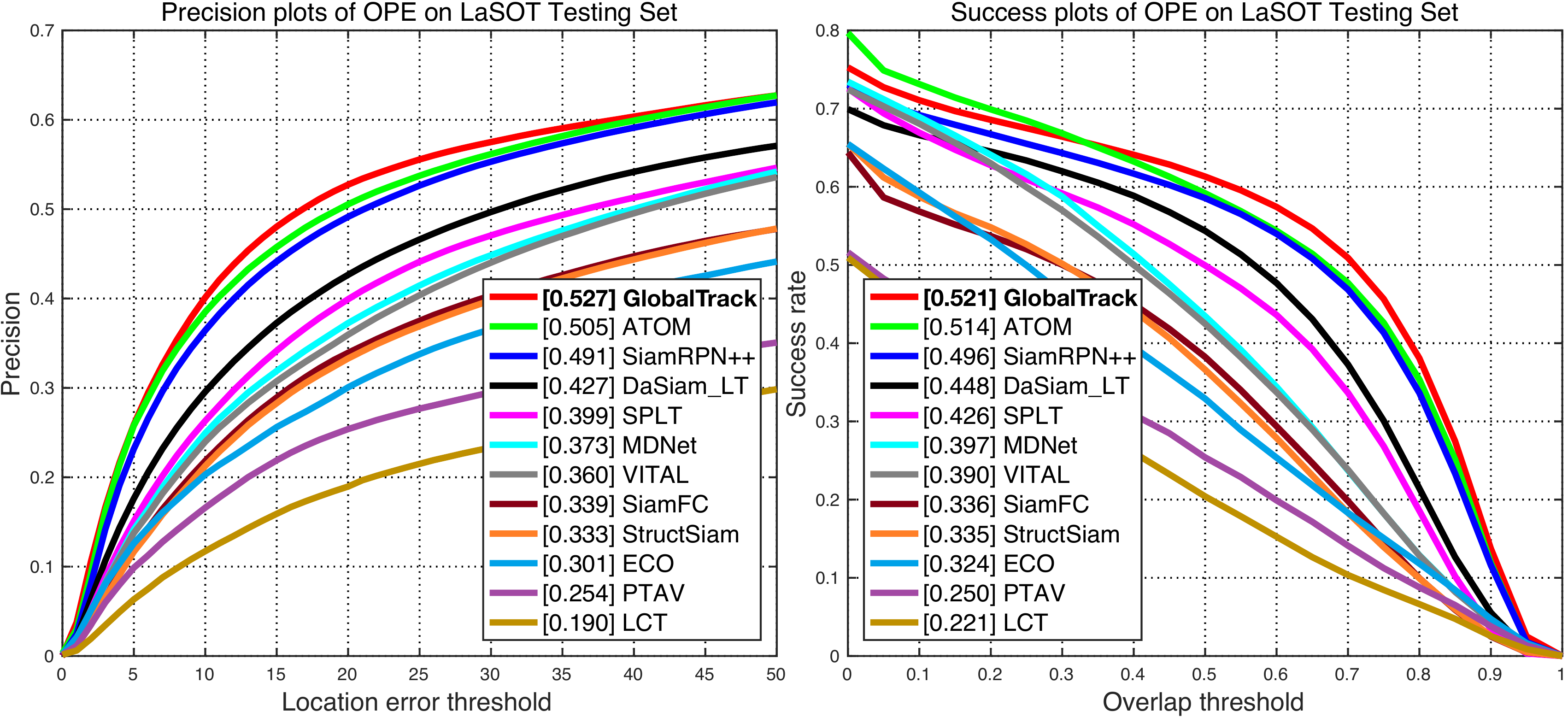}
  \caption{Success plots of GlobalTrack and state-of-the-art trackers on the test set of LaSOT. Compared to the previous best tracker ATOM, our approach achieves absolute gains of 2.2\% in precision score and 0.7\% in success score, respectively. Besides, GlobalTrack also outperforms the latest long-term trackers SPLT and DaSiam\_LT by a large margin.}
  \label{fig:lasot_results}
\end{figure*}

\subsection{Ablation Study}

In this section, we perform an extensive analysis of the proposed approach. Unless specified, the experiments are conducted on OTB-2015 dataset, which consists of 100 videos. We evaluate the approaches using overlap precision (OP), which indicates the percentage of successfully tracked frames where the overlap rates exceed $0.5$.

\subsubsection{Analysis of QG-RPN}
We compare the proposed query-guided RPN, which is able to adaptively change the predictions according to different queries, with the general purposed RPN~\cite{fasterrcnn2015} and its improved variant Guided-Anchor RPN (GA-RPN)~\cite{garpn2019}. The results are shown in Figure~\ref{fig:ar_k_rpns} and Table~\ref{tab:ar_k_rpns}. We use AR@k to evaluate these RPNs, which indicates the average recall using top-k predictions.
As shown in Figure~\ref{fig:ar_k_rpns}, QG-RPN consistently outperforms RPN and GA-RPN by a large margin, and it is able to achieve high recall with only a few proposals.
Table~\ref{tab:ar_k_rpns} shows the numerical comparison. Using only the top-1 prediction, QG-RPN achieves a recall of 67.1\%, surpassing RPN and GA-RPN by over 55\%.
With only the top-10 proposals, QG-RPN achieves a recall of 91.9\%, which is comparable with the top-500 recalls of RPN and GA-RPN, indicating the effeciency of our approach.
With top-2000 proposals, our approach also outperforms the compared approaches by achieving a recall of 99.2\%.
These results verify the effectiveness of QG-RPN and demonstrate its superiority in both recall rate and efficiency.

\begin{table}[t]
   \footnotesize
   \begin{center}
       \caption{State-of-the-art comparison on the long-term tracking benchmark TLP in terms of success rate (SR, under overlap threshold 0.5), success score and precision score.}
       \label{tab:tlp_results}
       \begin{tabular}{
            >{\raggedright\arraybackslash} m{2.1cm}
            >{\centering\arraybackslash} m{1.2cm}
            >{\centering\arraybackslash} m{1.2cm}
            >{\centering\arraybackslash} m{1.2cm}
            }
            \hline
            (\%)        & SR$_{0.5}$ & Success & Precision \\
            \hline
            LCT         & 8.7  & 9.9  & 7.2 \\
            ECO         & 21.9 & 20.2 & 21.2 \\
            ADNet       & 22.1 & 22.3 & 20.3 \\
            MDNet       & 42.3 & \textcolor{blue}{\bf 37.0} & \textcolor{blue}{\bf 38.4} \\
            MBMD        & 48.1 & - & - \\
            SiamRPN     & 51.5 & - & - \\
            ATOM        & 47.5 & - & - \\
            SPLT        & \textcolor{blue}{\bf 52.7} & - & - \\
            \hline
            {\bf GlobalTrack} & \textcolor{red}{\bf 63.8} & \textcolor{red}{\bf 52.0} & \textcolor{red}{\bf 55.6} \\
            \hline
       \end{tabular}
   \end{center}
\end{table}

\subsubsection{Comparison of QG-RPN and QG-RCNN}
QG-RPN and QG-RCNN are the two stages of our approach that pursue for recall and the top-1 accuracy, respectively.
We compare their AR@k results in Figure~\ref{fig:ar_k_rpn_rcnn} and Table~\ref{tab:ar_k_rpn_rcnn}.
As shown in Figure~\ref{fig:ar_k_rpn_rcnn}, the QG-RCNN achieves higher AR score than QG-RPN at the very start, and is soon surpassed by QG-RPN as more candidates are available.
From Table~\ref{tab:ar_k_rpn_rcnn} we observe that, the top-1 accuracy (which is exactly the AR@1) of QG-RCNN surpasses QG-RPN by up to 9.5\%. 
However, as more candidates are introduced, QG-RPN achieves significantly higher recall than QG-RCNN, even using only the top-5 predictions.
As $k$ gets even larger (e.g., $k \geq 100$), the ARs of QG-RPN and QG-RCNN become close.
The results verify the effectiveness of QG-RPN and QG-RCNN, and show their different preferences for accuracy and recall.

\subsubsection{Impact of Loss Function}
To verify the effectiveness of the proposed cross-query loss, we compare our model trained with cross-query loss and with single query loss -- the loss that only considers one instance per image pair.
We evaluate the two models on the test set of LaSOT dataset. The results are shown in Table~\ref{tab:loss_comparison}.
We observe that, the model trained with cross-query loss surpasses that trained with the single-query loss in all three metrics by $2.6\%\sim 4.2\%$, verifying the robustness of the proposed loss function.

\subsection{Comparison with State-of-the-art}

We compare our approach GlobalTrack with state-of-the-art trackers on four large-scale tracking benchmarks.
The compared approaches include SPLT~\cite{splt2019}, ATOM~\cite{atom2019}, SiamRPN++~\cite{siamrpnpp2019}, C-RPN~\cite{crpn19}, MBMD~\cite{mbmd2018}, DaSiam\_LT~\cite{dasiamrpn2018}, UPDT~\cite{updt2018}, VITAL~\cite{vital2018}, SINT~\cite{sint2016}, CF2~\cite{cf2_2016}, ADNet~\cite{adnet17}, MDNet~\cite{mdnet2016}, SiamFC~\cite{siamfc2016}, SiamFCv2~\cite{cfnet2017}, CFNet~\cite{cfnet2017}, StructSiam~\cite{structsiam2018}, ECO~\cite{eco2017}, PTAV~\cite{ptav2017}, TLD~\cite{tld2012} and LCT~\cite{lct2015}.

\begin{table}[t]
   \footnotesize
   \begin{center}
       \caption{State-of-the-art comparison on the test set of TrackingNet in terms of precision, normalized precision and success (AUC).}
       \label{tab:trackingnet_results}
       \begin{tabular}{
            >{\raggedright\arraybackslash} m{2.1cm}
            >{\centering\arraybackslash} m{1.1cm}
            >{\centering\arraybackslash} m{2.2cm}
            >{\centering\arraybackslash} m{1.1cm}
            }
            \hline
            (\%)        & Precision & Norm. Precision & Success \\
            \hline
            CFNet       & 53.3 & 65.4 & 57.8 \\
            MDNet       & 56.5 & 70.5 & 60.6 \\
            ECO         & 49.2 & 61.8 & 55.4 \\
            UPDT        & 55.7 & 70.2 & 61.1 \\
            DaSiamRPN   & 59.1 & 73.3 & 63.8 \\
            SiamRPN++   & \textcolor{red}{\bf 69.4} & \textcolor{red}{\bf 80.0} & \textcolor{red}{\bf 73.3} \\
            C-RPN       & 61.9 & 74.6 & 66.9 \\
            ATOM        & 64.8 & \textcolor{blue}{\bf 77.1} & 70.3 \\
            \hline
            {\bf GlobalTrack} & \textcolor{blue}{\bf 65.6} & 75.4 & \textcolor{blue}{\bf 70.4} \\
            \hline
       \end{tabular}
   \end{center}
\end{table}

\subsubsection{LaSOT}
The test set of LaSOT consists of 280 videos with an average length of 2448 frames, which is longer than most other datasets. The precision and success plots on LaSOT are shown in Figure~\ref{fig:lasot_results}.
Compared to the previous best approaches ATOM and SiamRPN++, our approach achieves an absolute gain of 2.2\% and 3.6\% in precision score, and 0.7\% and 2.5\% in success score, respectively.
The tracking results demonstrate the advantage of global instance search in long-term tracking.

\subsubsection{TLP}
The dataset consists of 50 long videos with an average length of 13529 frames, which is much longer than all other tracking datasets.
The results of the trackers are shown in Table~\ref{tab:tlp_results}.
Our approach GlobalTrack outperforms all other trackers, including the latest long-term tracker SPLT, by a very large margin ($\sim 11.1\%$ absolute gain in SR$_{0.5}$), verifying the significant advantage of our approach over all existing algorithms in tracking very long videos.



\begin{table}[t]
   \footnotesize
   \begin{center}
       \caption{State-of-the-art comparison on the {\bf test set} of OxUvA in terms of maximum geometric mean (MaxGM), true positve rate (TPR) and true negative rate (TNR).}
       \label{tab:oxuva_results}
       \begin{tabular}{
            >{\raggedright\arraybackslash} m{2.1cm}
            >{\centering\arraybackslash} m{1.2cm}
            >{\centering\arraybackslash} m{1.2cm}
            >{\centering\arraybackslash} m{1.2cm}
            }
            \hline
            (\%)        & MaxGM & TPR & TNR \\
            \hline
            ECO-HC      & 31.4 & 39.5 & 0.0 \\
            MDNet       & 34.3 & 47.2 & 0.0 \\
            TLD         & 43.1 & 20.8 & \textcolor{red}{\bf 89.5} \\
            LCT         & 39.6 & 29.2 & 53.7 \\
            SINT        & 32.6 & 42.6 & 0.0 \\
            SiamFC+R    & 45.4 & 42.7 & 48.1 \\
            EBT         & 28.3 & 32.1 & 0.0 \\
            DaSiam\_LT  & 41.5 & 68.9 & 0.0 \\
            MBMD        & 54.4 & 60.9 & 48.5 \\
            SPLT        & \textcolor{red}{\bf 62.2} & \textcolor{blue}{\bf 49.8} & \textcolor{blue}{\bf 77.6} \\
            \hline
            {\bf GlobalTrack} & \textcolor{blue}{\bf 60.3} & \textcolor{red}{\bf 57.4} & 63.3 \\
            \hline
       \end{tabular}
   \end{center}
\end{table}

\subsubsection{TrackingNet}
The test set of TrackingNet consists of 511 videos collected from the YouTube website.
The evaluation results of the trackers are shown in Table~\ref{tab:trackingnet_results}.
SiamRPN++ and ATOM achieve impressive success scores of 73.3\% and 70.3\%, respectively, while our approach achieves a success score of 70.4\%, which is comparable with the best trackers.
The tracking results show the generalization ability of our approach on large test data.

\subsubsection{OxUvA}
OxUvA is a long-term tracking benchmark, where its development and test sets consist of 200 and 166 videos, respectively. The average video length of OxUvA is 4260 frames, which is much longer than most other datasets.
The protocol of OxUvA requires the approach to submit not only the estimated bounding boxes, but also the prediction of target absences in all frames.
To generate the presence/absence prediction, we simply threshold the top-1 scores of QG-RCNN, where the frames with top-1 scores exceed $\tau=0.84$ are considered \textit{target presence}, while those with top-1 scores below the threshold are considered \textit{target absence}.
The evaluation results of the trackers on the test and development sets of OxUvA are shown in Table~\ref{tab:oxuva_results} and Table~\ref{tab:oxuva_dev_results}, respectively.
Compared to the reported best tracker SiamFC-R, our approach achieves an absolute gain of 14.9\% and 24.2\% on the test and development sets of OxUvA in terms of MaxGM (maximum geometric mean of TPR and TNR), respectively.
The results demonstrate the strong performance of our approach in long-term tracking scenarios.

\section{Conclusions and Future Works}

In this work, we propose a pure global instance search based tracker that imposes no assumption or constraint on temporal consistency. Therefore, its performance in a frame is not affected by previous tracking failures, which makes it ideal for long-term tracking. The method is developed based on two-stage object detectors, and it is composed of two components: a query-guided region proposal network for generating query-specific instance candidates, and a query-guided region convolutional neural network for classifying these candidates and generating the final predictions. Experiments on four large-scale tracking benchmarks verify the strong performance of the proposed approach.

\begin{table}[t]
   \footnotesize
   \begin{center}
       \caption{State-of-the-art comparison on the {\bf development set} of OxUvA in terms of maximum geometric mean (MaxGM), true positve rate (TPR) and true negative rate (TNR).}
       \label{tab:oxuva_dev_results}
       \begin{tabular}{
            >{\raggedright\arraybackslash} m{2.1cm}
            >{\centering\arraybackslash} m{1.2cm}
            >{\centering\arraybackslash} m{1.2cm}
            >{\centering\arraybackslash} m{1.2cm}
            }
            \hline
            (\%)        & MaxGM & TPR & TNR \\
            \hline
            ECO-HC      & 26.6 & 28.3 & 0.0 \\
            MDNet       & 32.4 & \textcolor{blue}{\bf 42.1} & 0.0 \\
            TLD         & 36.6 & 14.1 & \textcolor{red}{\bf 94.9} \\
            LCT         & 31.7 & 22.7 & 43.2 \\
            SINT        & 30.3 & 26.8 & 0.0 \\
            SiamFC+R    & \textcolor{blue}{\bf 39.7} & 35.4 & 43.8 \\
            \hline
            {\bf GlobalTrack} & \textcolor{red}{\bf 63.9} & \textcolor{red}{\bf 55.8} & \textcolor{blue}{\bf 73.2} \\
            \hline
       \end{tabular}
   \end{center}
\end{table}


\section{ Acknowledgments}
This work is supported in part by the National Key Research and Development Program of China (Grant No. 2016YFB1001005), the National Natural Science Foundation of China (Grant No. 61602485 and No. 61673375), and the Projects of Chinese Academy of Science (Grant No. QYZDB-SSW-JSC006).

\bibliographystyle{aaai}
\bibliography{AAAI-HuangL.7322.bib}

\end{document}